\documentclass[a4paper,fleqn]{cas-sc}

\usepackage[authoryear,longnamesfirst]{natbib}

\usepackage[polutonikogreek,frenchb]{babel}
\usepackage{xcolor}
\usepackage{comment}

\def\tsc#1{\csdef{#1}{\textsc{\lowercase{#1}}\xspace}}
\tsc{WGM}
\tsc{QE}

\begin{document}
\let\WriteBookmarks\relax
\def\floatpagepagefraction{1}
\def\textpagefraction{.001}

\shorttitle{Character Recognition in Byzantine Seals}
\shortauthors{Rageau \& al.}  

\title [mode = title]{Character Recognition in Byzantine Seals  with Deep Neural Networks}                      
\tnotemark[1]
\tnotemark[<tnote number>] 
\tnotetext[1]{This document is the result of the research
   project funded by the ANR National  Research Agency, grant number ANR-21-CE38-0001, \url{https://anr.fr/Projet-ANR-21-CE38-0001}}


%




\ead{email address}

\ead[url]{URL}


\author[1,6]{Théophile Rageau}

\ead{theophilerageau041@gmail.com}


\credit{Conceptualization of this study, Methodology, Software}

\author[1]{Laurence Likforman-Sulem}[orcid=0000-0001-9096-7239]
\cormark[2]
\ead{likforman@telecom-paris.fr}
\author[1,5]{Attilio Fiandrotti}[orcid=0000-0002-9991-6822]
\ead{attilio.fiandrotti@unito.it}
\credit{Data curation, Writing - Original draft preparation}
\author[2]{Victoria Eyharabide}[orcid=0000-0002-3775-1495]
\ead{maria-victoria.eyharabide@sorbonne-universite.fr}
\author[3]{Béatrice Caseau}[orcid=0000-0002-6308-1261]
\ead{beatrice.caseau@sorbonne-universite.fr}

\author[4]{Jean-Claude Cheynet}
\ead{jean-claude.cheynet@college-de-france.fr}

\cortext[cor1]{L. Likforman-Sulem}

\affiliation[1]{organization={IDS Department, LTCI, Institut Polytechnique de Paris/Telecom Paris},
    addressline={19 Place Marguerite Perey}, 
    city={Palaiseau},
    postcode={911231}, 
    country={France}}
\affiliation[2]{organization={STIH Laboratory, Sorbonne University},
    addressline={28 rue Serpente}, 
    city={Paris},
    postcode={75006}, 
    country={France}}

\affiliation[3]{organization={UMR 8167 Orient \& Méditerranée, Sorbonne University},
    addressline={1 rue Victor Cousin}, 
    city={Paris},
    postcode={75005}, 
    country={France}}

\affiliation[4]{organization={Collège de France, CNRS},
    addressline={52 rue du Cardinal Lemoine}, 
    city={Paris},
    postcode={75231}, 
    country={France}}

\affiliation[5]{organization={Universit\'a di Torino, UniTo},
    addressline={12 via Pessinetto}, 
    city={Torino},
    postcode={10153}, 
    country={Italy}}

\affiliation[6]{organization={École normale supérieure Paris-Saclay, Université Paris-Saclay},
    addressline={4 Avenue des Sciences}, 
    city={Gif-sur-Yvette},
    postcode={91190}, 
    country={France}}








\begin{abstract}
Seals are small coin-shaped artifacts, mostly made of lead, held with strings to seal letters.
This work presents the first attempt towards automatic reading of text on Byzantine seal images.
Byzantine seals are generally decorated with iconography on the obverse side and Greek text on the reverse side. Text may include the sender's name, position in the Byzantine aristocracy, and elements of prayers.
Both text and iconography are precious literary sources that wait to be exploited electronically, so the development of computerized systems for interpreting seals images is of paramount importance.
This work's contribution is hence a deep, two-stages, character reading pipeline for transcribing Byzantine seal images. A first deep convolutional neural network (CNN) detects characters in the seal (character localization).
A second convolutional network reads the localized characters (character classification). 
Finally, a diplomatic transcription of the seal is provided by post-processing the two network outputs.
We provide an experimental evaluation of each CNN in isolation and both CNNs in combination. All performances are evaluated by cross-validation. Character localization achieves a mean average precision (mAP@0.5) greater than 0.9. Classification of characters cropped from ground truth bounding boxes achieves Top-1 accuracy greater than 0.92.  End-to-end evaluation shows the efficiency of the proposed approach when compared to the SoTA for similar tasks.
\end{abstract}


\begin{highlights}
\item We present the first system dedicated to the recognition of characters in Byzantine seal images.
\item A two-stages deep architecture is proposed for localizing and classifying characters. 
\item A character-level diplomatic transcription is then obtained (by reordering characters into text lines) as the first step towards language processing to extract words, named entities and recover the underlying text.
\end{highlights}


\maketitle

\section{Introduction}\label{sec:intro}

Byzantine sigillography is the study of seals that were produced during the Byzantine period. Seals are small circular artifacts (made of lead or sometimes gold) that were held with a string to 'seal' official or private documents or objects. Mentions of family names, function in the State and dignities provide important information on the aristocracy and the administration of the Byzantine State. In this article we focus on Byzantine seals from the VIII-XII\textsuperscript{th} centuries. They often contain the image of a saint or a cross on the obverse side and provide information on the preferred religious motif of the sigillant (\cite{Zacos, Tatish}). More than 60,000 seals have been unearthed and are preserved in public or private collections worldwide. 

Seals have two sides: obverse and reverse. On the obverse side, one often finds iconography. On the reverse side, a text is usually written in Byzantine Greek capital letters, including three to seven 'text lines' depending on the size of the seal and of the letters. This text may contain the name of the sender, his/her social position, and elements of prayers. Sometimes, either side or both sides may contain a monogram that the document's reader should decipher by combining letters.
Text on seals is composed of ancient Greek letters, whose glyphs have evolved over time. These variations are encoded by the Athena Ruby font (\cite{Kalve2015}). This font was conceived specially for Byzantine seals and includes about 700 distinct glyphs.

\begin{figure}[h!]
    \centering
    \includegraphics[width=0.6\linewidth]{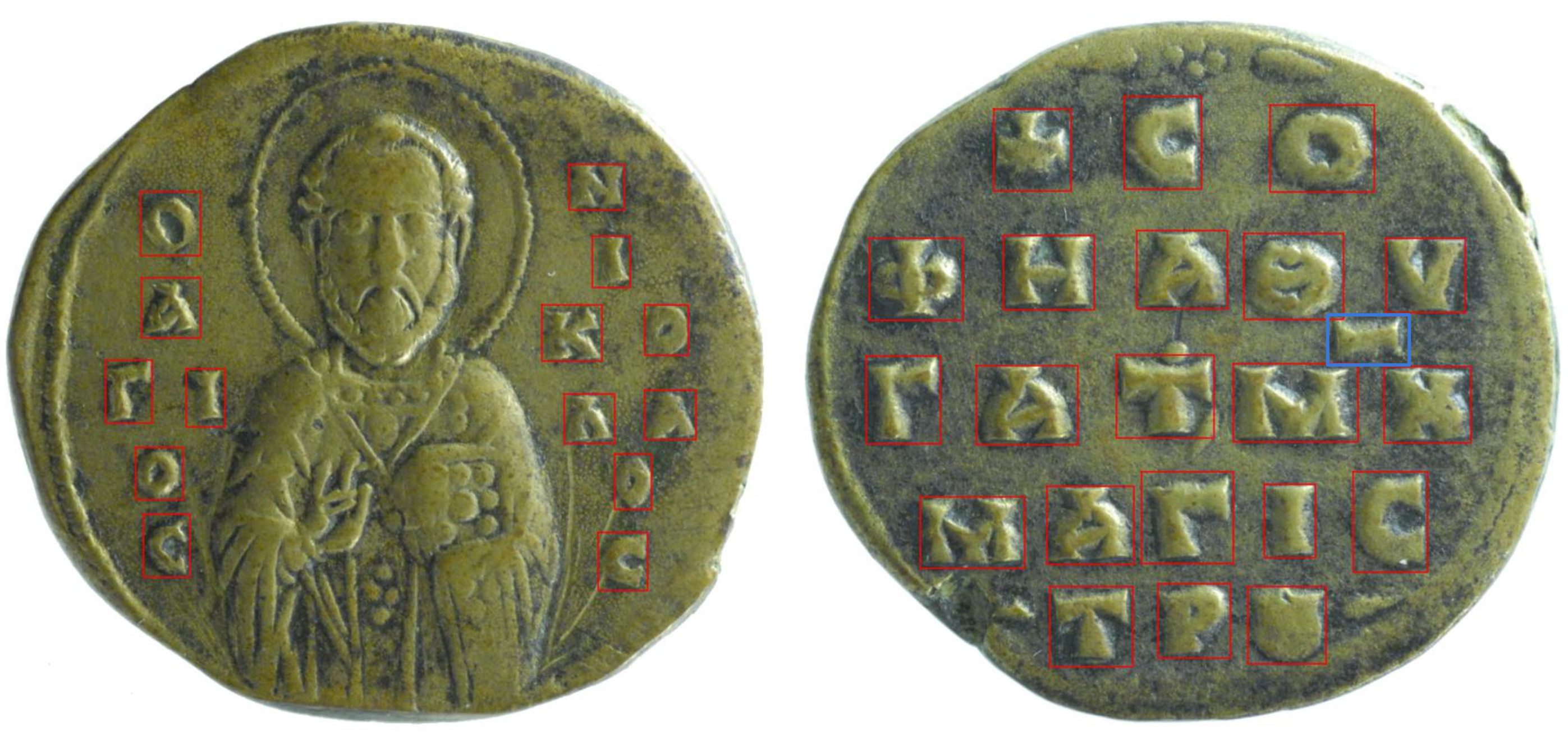}
    \caption{Sample seal images. The obverse side (left) includes iconography. The reverse side (right) includes text of same color as  background. Ground truth character bounding boxes are superimposed in red color. An abbreviation marker (right) is superimposed in blue color.}
    \label{fig:exemple_seals}
\end{figure}

\begin{figure}[h!]
    \centering
    \includegraphics[width=0.6\linewidth]{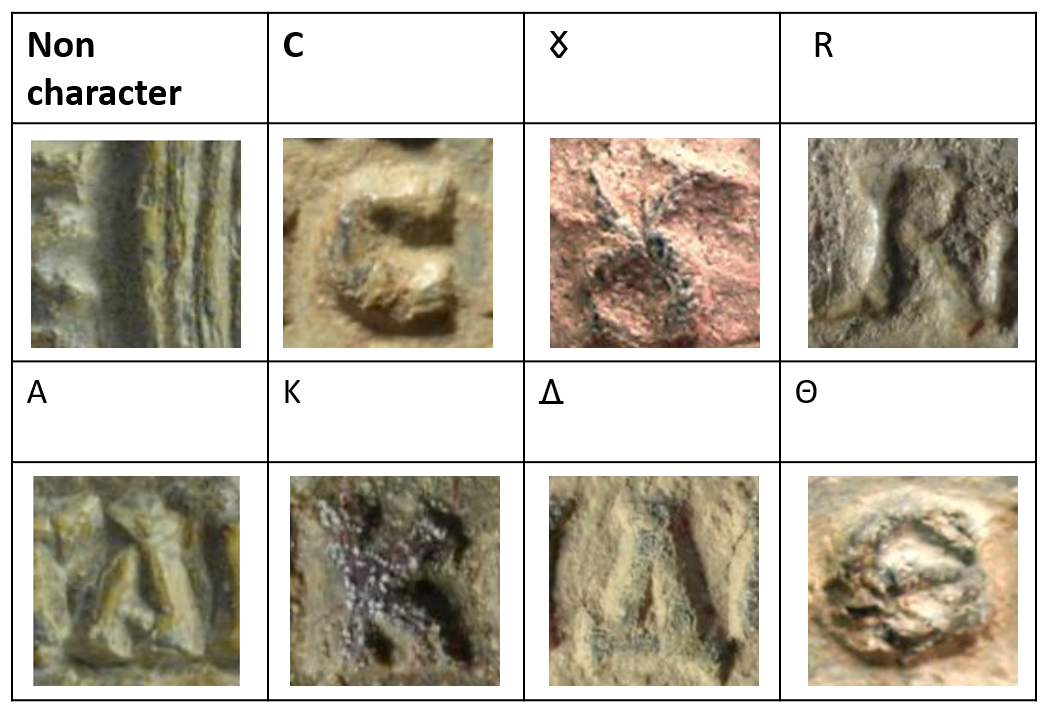}
    \caption{Sample character images cropped from seal images. Such images are challenging for character recognition. }
    \label{fig:sample_chars}
\end{figure}

Figure \ref{fig:exemple_seals} shows the obverse and reverse sides of a sample seal. The characters are made by an engraver who creates reliefs on a \textit{boulloterion}, the tool used to strike lead, silver and gold bullae. Digitization has been performed by a camera. Notice that  background and  characters have the same color, and in addition shades may be present depending on the light source position during the digitization process. Therefore, traditional OCR approaches cannot be applied to such images since they require contrast between foreground (characters) and background (see Fig. \ref{fig:sample_chars}) 
%

The scene depicted in this seal (left) is St. Nicholas in his customary episcopal garb, holding the Book of the Gospels in his left hand and blessing with his right.

Since the surface of a seal is small (between 10 to 50 mm in diameter), engravers have gained room by removing word spaces, omitting or fusing characters, and omitting even entire words. Consequently, the text is often abbreviated.
As an example: 
the inscription on the reverse side of the seal presented in Figure \ref{fig:exemple_seals} \textgreek{+Σοφήα θυγάτ Μχ μαγίστρου+}
actually corresponds to a longer text that starts with the "croisette" symbol  (unseen characters are boxed): 
\textgreek{+Σοφήα θυγάτ\framebox[1.1\width]{ηρ} Μ\framebox[1.1\width]{ι}χ\framebox[1.1\width]{αὴλ} μαγίστρου+},
Sofía thygátir Michaíl magístrou (in Greek), which means: "Sophie, daughter of  Michel magistros" (English translation).
%
%
\begin{figure}[h!]
    \centering
    \includegraphics[width=0.5\linewidth]{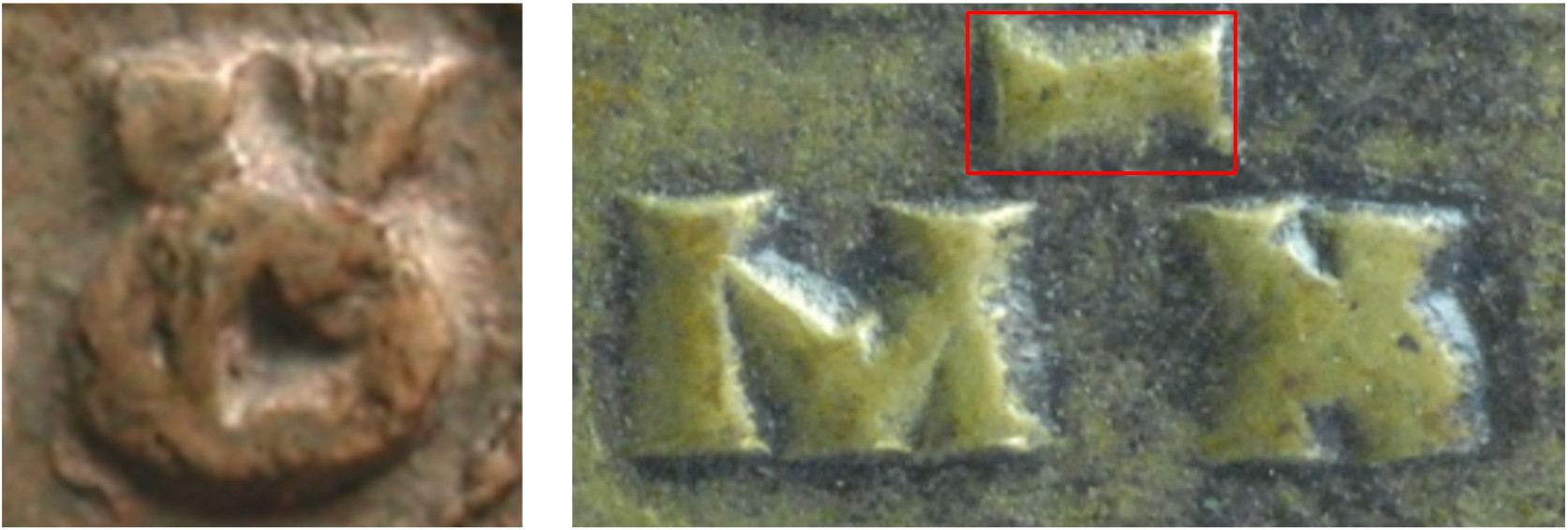}
    \caption{Examples of text abbreviation.  On the left ligature $o \upsilon$, on the right the upper mark symbolizes an abbreviated form: $M \chi$ corresponds to  word $M \iota \chi \alpha \eta \lambda$ (Miha\"il or Michael).}
    \label{fig:ligatures}
\end{figure}

Figure \ref{fig:ligatures} illustrates this abbreviation process in Byzantine seals. The first one corresponds to the ligature  "$o \upsilon$" (two characters fused into one symbol). The pattern  "$o \upsilon$" corresponds to a widespread grammatical form in Greek. The second one corresponds to one word (Michael), reduced to its first (M) and third ($\chi$) letters.  

\begin{figure*}[h!]
    \centering
    \includegraphics[width=0.9\linewidth]{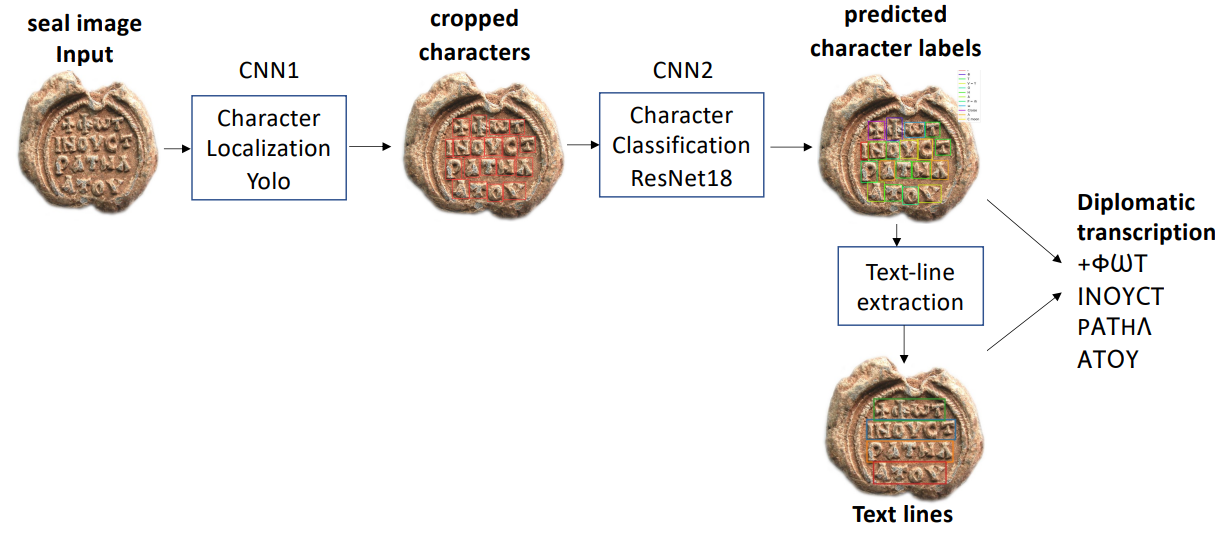}
    \caption{Pipeline of the proposed two-stage character recognition approach. A YOLO network detects characters within each seal providing bounding boxes (object localization). Next, a ResNet18 network classifies each detected character (object classification). The output is the diplomatic transcription of the input reverse seal image. }
    \label{fig:bhai}
\end{figure*}

To help historians interpret seals, the \textit{BHAI}  project 
proposes an innovative approach for Byzantine sigillography based on artificial intelligence \footnote{This  project is lead by V. Eyharabide, in collaboration with historians, experts in Byzantine history and sigillography (Prof. J-C Cheynet and B. Caseau, L. Orlandi, and A. Binoux) \url{https://anr.fr/Projet-ANR-21-CE38-0001}
}.
The objectives of this project include seal dating, iconography interpretation (what the scene represents), and text recovery (despite its abbreviated form and damaged characters). These objectives rely on recognition tasks such as figures and objects recognition (obverse) and character recognition (reverse).
This paper proposes a learnable two-stage character reading approach capable of producing a diplomatic transcription of seal images (Fig. \ref{fig:bhai}). Our approach is implemented as a pipeline that includes the following stages:

\begin{itemize}
\item a convolutional object detector trained to detect  characters from seal images (character localization). From the predicted bounding boxes, isolated character images can be cropped.
\item a convolutional image classifier trained to recognize letters in the crops above  (character classification)
  \item finding the text lines and transcribing reverse seal images from characters' positions and labels.
\end{itemize}
To our knowledge, the proposed system is the first recognition system dedicated to the recognition of seal images, in particular for Byzantine seals. Our  approach has been developed in the context of low resource data, since annotated seal images are scarce.
%
%

The rest of the paper is organized as follows. We present related works dealing with cultural heritage and written artifacts in Section \ref{sec_related}. In Section \ref{sec:dataset}, we describe our annotated dataset of Byzantine seals from two different collections. Our proposed two-stages approach is detailed in Section \ref{ObjectDetection}, including their architectures (Sections \ref{sec:localization} and \ref{sec:Resnet}, respectively) and training procedures. Finally, in Section \ref{sec:experiments} we experiment at reading characters from our dataset.

\section{Related works}
\label{sec_related}
Ancient written documentary sources are important for understanding the history of a period of time in a specific geographical area. Such sources were written for the most ancient ones on argile and wood tablets (\cite{Rusakov21}, \cite{Molton02}), papyri (\cite{Pirrone2021}),  parchments (\cite{DhaliJWS20}, \cite{eglin14}),  funerary inscriptions and funerary stele (\cite{BOBOU2020e00164,ANDREU2019e00091}), metal especially coins (\cite{AnCoins}), and stones (\cite{Sfikas20}). Conferences such as DATeCH (\cite{datech_proc}) highlight the interest of recognizing the text content of ancient documents. 

These last decades, several projects have been launched in order to build computerized systems for automatically deciphering digitized written artifacts (ILAC for coins, \cite{AncientRomanCoins}, ARCADIA for pottery sherds (\cite{CHETOUANI20201}). 
%
However, collections may be hard to interpret even for human beings, mainly due to their damaged state and the lack of ancient language knowledge and historical context (\cite{Souibgui022}).
The above systems rely on machine learning approaches. Such approaches require a  training step that uses annotated data. 
In the best cases, domain experts can annotate digitized written material to provide transcriptions, meta-data, and bounding boxes of objects of interest. Nevertheless, annotation is a time-consuming task requiring expertise and knowledge. Therefore, only a limited number of images may be annotated.

Depending on the manufacturing process, it will be more or less easy for a computer to read written artifacts. In particular, well-contrasted writings with separated characters are the easiest to recognize. Moreover, writings on coins, tablets, and seals result from a 3-dimensional manufacturing process. Digitization into 2-dimensional images with a camera provides shadows (depending on the light source and contrast) that are difficult to process. An additional difficulty is the size of the letters and the style of writing.

Scene Text recognition approaches aim to read the text in natural scenes images digitized by a camera (\cite{bai16}). Such approaches may also be suitable for other written artifacts. In this context, there are two main approaches: one based on machine learning and the other on deep learning (a mix of both is also possible when several tasks are required).  
The machine learning approach consists in extracting features (f.i. SIFT features) from keypoints, then defining a matching or classifying based approach  (f.i. SVMs). Such approach  is described in (\cite{AncientRomanCoins}) for classifying coins into 65 classes (portrayed Emperors). But there is no attempt to read  characters.  In contrast \cite{Kavelar14}, aims at   reading Emperor names from recognized characters. Depending on the lexicon and other parameters, the Word Recognition Rate ranges from 29 to  67\%. The objective in  (\cite{pan17}) is to read  coin release dates from recognized digits.  
Average recognition accuracy is about 80\% for digit images collected under well controlled lighting conditions.

\begin{figure}[b!]
    \centering
    \includegraphics[width=0.6\linewidth]{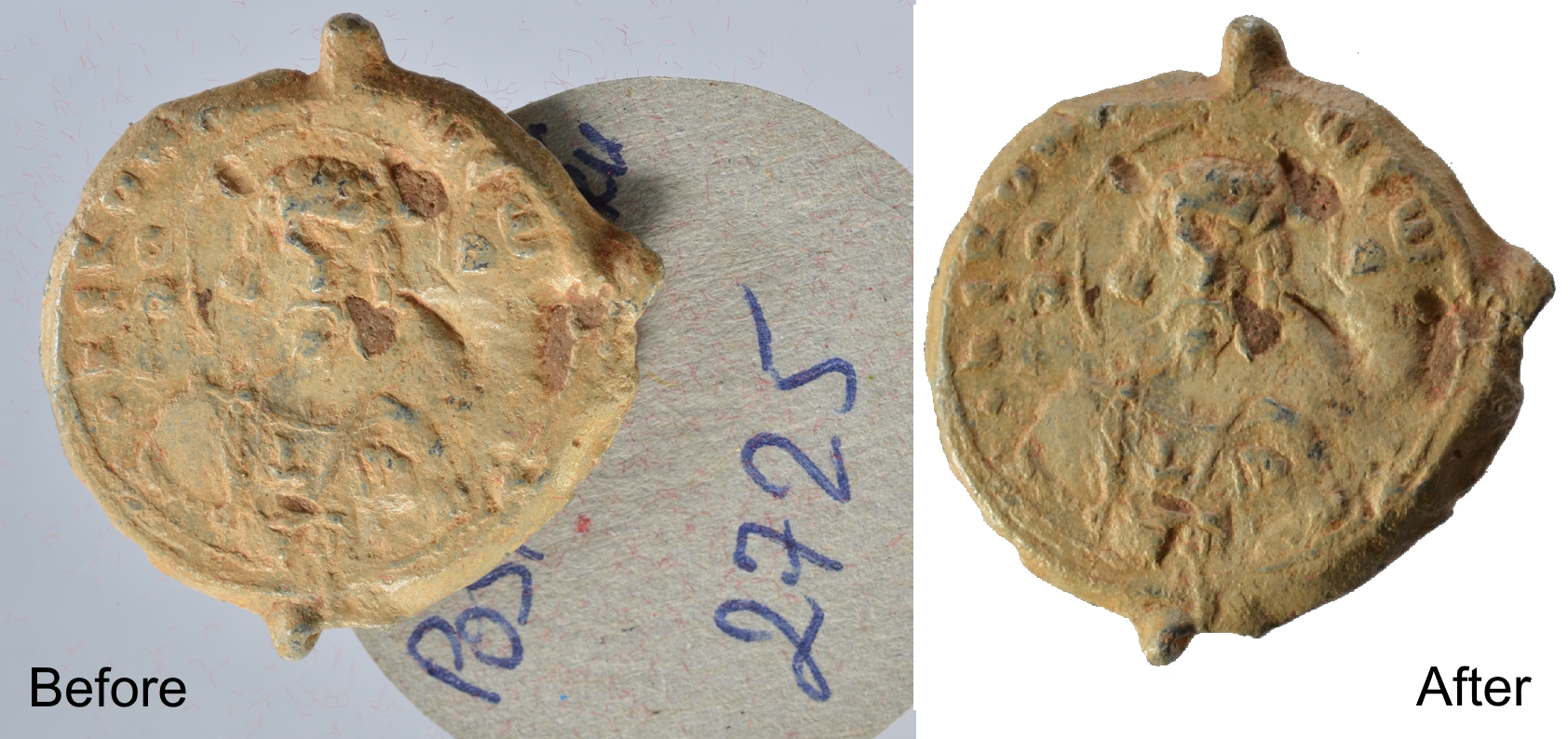}
    \caption{Example of a seal image before and after preprocessing}
    \label{fig:BeforeAfter}
\end{figure}

In contrast to Machine Learning approaches, deep learning approaches automatically learn features and store them in their convolutional layers. If enough annotated data are available for training, it is possible to train multi-class systems that both localize and recognize textual elements (\cite{bartz17}). However, data scarcity is a common problem for ancient artifacts. One solution is the generation of synthetic data (\cite{elsana22}) and fine-tuning on real data (\cite{Nguyen21}). As a result, millions of synthetic characters may be necessary. 
Another approach consists of first locating text regions and then applying a character recognition approach in each text region. These regions can enclose a paragraph, a text line, or a character. This two-step approach is suitable for facing data scarcity: the combined localization/recognition task is decomposed into two more straightforward tasks. As a result, two networks are created, each being lighter than a single large network. 

In the present work, we face data scarcity, low-contrast images, and damaged characters. Thus, we decompose the recognition task into two stages:  i) character localization and ii) character classification (see Sections \ref{sec:localization} and \ref{sec:Resnet}).
%
%
%
%

\section{Seal image dataset}
\label{sec:dataset}

 Our dataset includes seal images from two collections of Byzantine seals: the Geneva Zacos collection and the Tatish collection. George Zacos collected the seals of the first collection, and her widow Janet Zacos gave a number of those seals to the "Musée d'art et d'histoire de Genève" in Switzerland. The second collection belongs to Nuri Yavuz Tatiş, and it is one of the richest and finest private seals collections in Turkey.

%
As shown in Fig. \ref{fig:BeforeAfter}, all images were manually cropped to center the image on the seal and eliminate all irrelevant surrounding information, such as handwritten notes. The background in each image was set to white to eliminate any side default since the seals had been photographed originally on different supports. In some cases, the images were also rotated to correctly display characters and iconographic scenes 
Their size is 1000$\times$1000 pixels (high resolution). 

\begin{figure}[h!]
    \centering
    \includegraphics[width=0.3\linewidth]{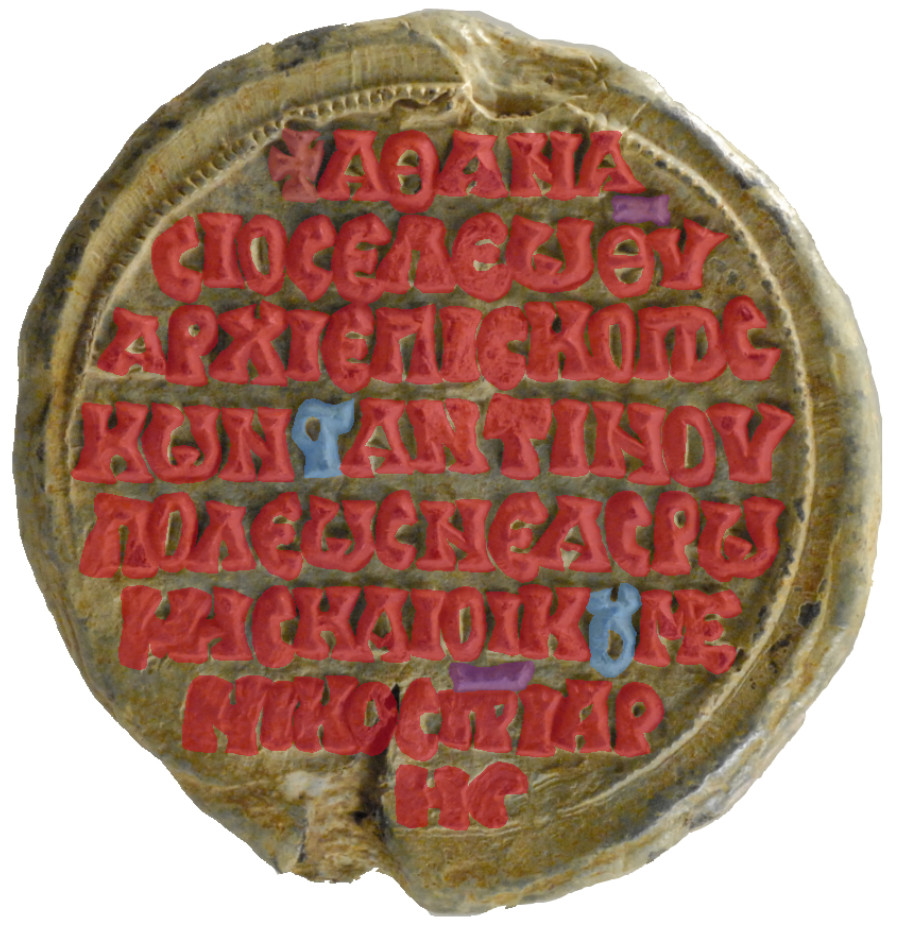} %
    \caption{Annotated seal at pixel level (reverse side). Character color codes: red=character, blue=ligature, 
    purple=abbreviation mark. Characters are in capital letters and  there is no space between words.}
    \label{fig:annotations_seals}
\end{figure}

Thanks to trained \textit{Anonymous}  experts in Greek and Byzantine history, characters and objects on seals of the Geneva Zacos collection were manually segmented at pixel level (semantic segmentation). 
They also provided a transcription in a standard Greek font and in the specialized Athena Ruby font (see Fig. \ref{fig:annotations_seals}).
%
%


\begin{figure}[h!]
    \centering
    \includegraphics[width=8cm]{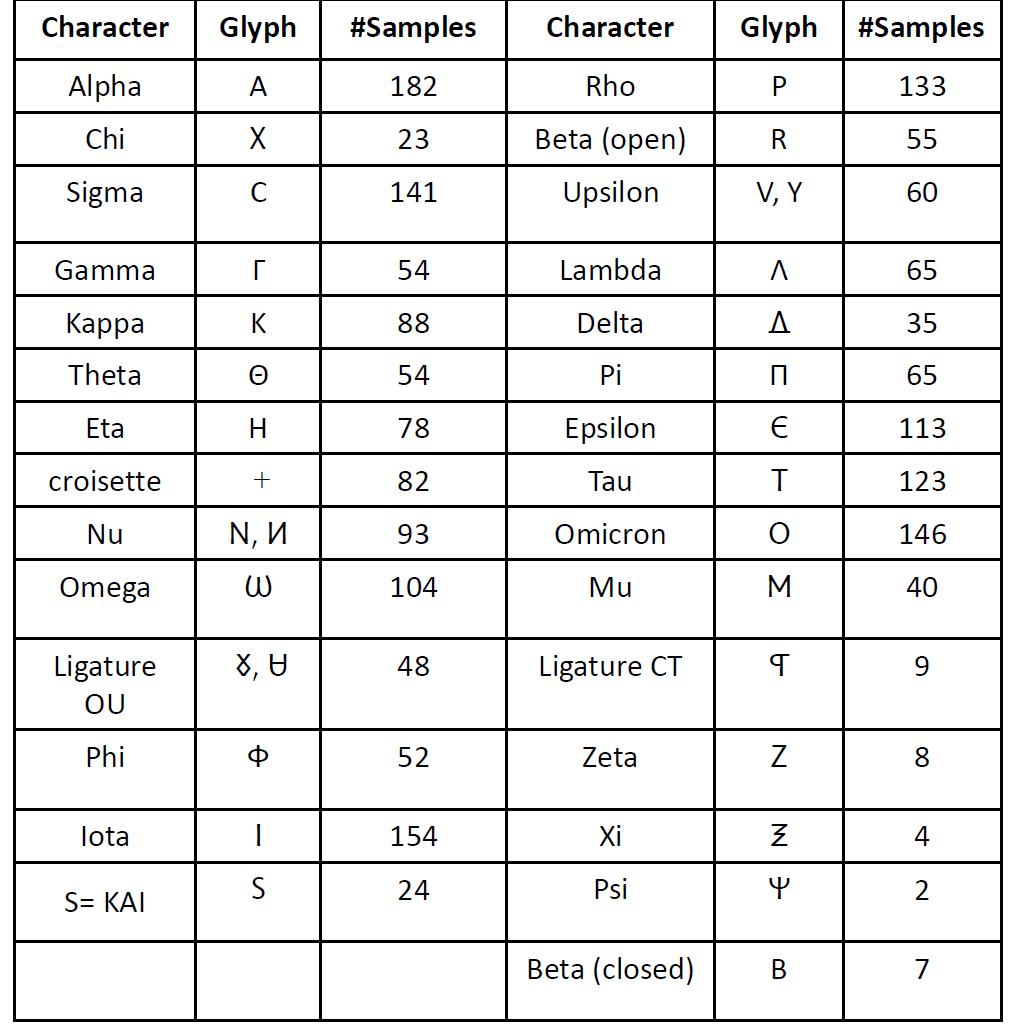}
    \caption{Ancient Greek characters with glyphs and their occurrences. The last 5 characters have fewer than 10 samples and were not considered in our experiments.}
    \label{fig:gliphs}
\end{figure}
%
Figure \ref{fig:gliphs} shows the 29 character classes we use in this work. Classes $\xi, \psi, \zeta$ and closed $\beta$  are quite unfrequent. The Upsilon and Nu classes include two different glyphs each. 
Character S is the abbreviated form of  word \textgreek{Kai}. There are also two ligatures: CT for \textgreek{st} (unfrequent) and OU for \textgreek{ou}.
%
When transcribed in standard Greek, the number of character classes drops to 
24 classes, the number of characters in the Greek alphabet.

From the pixel-based annotations, a bounding box around each character was easily derived. The number of seals we kept in the Geneva Zacos collection is 57 (29 reverse, 28 obverse).
The Tatish collection was not annotated at pixel level. Only transcriptions in a standard Greek font were provided. Thus we manually annotated 37 additional reverse and 8 obverse seal images by drawing bounding boxes around characters and assigning them a label. Such annotation is less complete compared to the Geneva Zacos collection however.
%
A total of 102 annotated images from both collections are available to train and evaluate the proposed approach (Table \ref{tab:nb_images}).

\begin{table}[]
    \caption{Number of annotated seal images. The annotation includes at least bounding boxes coordinates and character classes.}
    \centering
    \begin{tabular}{||c|c c c||}
    \hline
      Collection   &  \#reverse  & \#obverse  & \#total seal images\\
      \hline
      Zacos & 29 & 28 & 57\\
      \hline
      Tatish & 37 & 8 & 45 \\
      \hline
      Total & 66 & 36 & 102 \\
      \hline
    \end{tabular}
    \label{tab:nb_images}
\end{table}

\section{Proposed Approach}
\label{ObjectDetection}

In this section, we describe our learnable approach in order to transcribe text present on seal images. 
Speaking of text detection in general, two approaches are possible. The first and most desirable approach consists in jointly detecting and classifying the characters. This approach is known as \emph{object detection in the wild} and has been solved before for several different tasks.
For example, YOLO is a family of one-shot joint object detection and classification deep convolutional architectures trainable end-to-end in a single shot.
However, even if properly pretrained, such approaches usually require large annotated training sets in the order of tens of thousands of images, whereas we have far fewer annotated images in our case.
Therefore, we opted for a different approach consisting of i) localizing character bounding boxes in the image and ii) reading out the characters previously localized as a simple image classification approach. The advantage of this second approach is that the detector needs to solve a simpler task (just localization) instead of two tasks simultaneously (localization and recognition). This approach splits a larger problem into two inherently simpler sub-problems, where each can be solved by learnable models trained over far fewer annotated samples. In the conclusion section, we discuss possible ways to overcome this issue.

%
%
Finally, diplomatic transcriptions are produced combining the outputs of both networks and apply a Hough-based approach that sorts the detected and classified characters into text lines.

\subsection{Character localization}\label{sec:localization}

As a first step, we rely on the \textit{small} version of YOLO v5 (\cite{Yolov5}) to detect characters in the image (object localization).
The output of YOLO is a list of tuples for each object detected in the image. Each tuple has the form (x, y, width, height), where (x, y) stands for the re-normalized coordinates of the center of the predicted bounding box and (width, height) corresponds to the re-normalized size of the bounding box. Namely, our YOLO learnable parameters have been previously initialized by training the network over the COCO dataset (328,000 images). We refine such a network by fine-tuning the parameters over  seal images for 300 epochs.
%
%
Due to the limited number of data, we massively resort to geometric transformation (image shifts, scale variations) to augment the training set diversity. We optimize network parameters (i.e. network weights) with SGD (Stochastic Gradient Descent) for 300 epochs (the apparently large epoch count is due to the limited size of our actual training set) and a linear learning rate scheduler from 0.01 to 0.00001.

We experimented with training batches of 2, 4, 8, and 16 images, and the best performance-complexity tradeoff was obtained with a size equal to 4. At the end of each epoch, the model is evaluated on a validation dataset (15\% of the training  data) and the mAP mean average precision and F1-score are computed \cite{Padilla20}. The best model is then evaluated against testing data, i.e., a set of images the network has not seen at training time.


\subsection{Character classification}\label{sec:Resnet}
For character classification, we rely on a ResNet18 architecture (\cite{resnet}) pre-trained on the ImageNet dataset (\cite{ImageNet}), recasting the character reading problem as a crop classification task. We performed a number of preliminary experiments concluding that this architecture offers good enough classification accuracy even when trained over a limited number of annotated samples. The network is trained over isolated character crops extracted from the training images resized to 256$\times$256 pixels. To cope with data scarcity, we train the classifier on reverse and obverse   character images. In addition, we rely on augmentation strategies to increment the apparent training set size.
The number of character classes in our setup is 24, corresponding to the most frequent classes in Figure \ref{fig:gliphs}. 
The Croisette class was included (even if it is not a letter) to avoid confusion since its shape is similar to a character.

While the ResNet18 is trained over character-exact crops from the annotated training images, it is expected to operate on crops extracted by the YOLO-based localization pipeline when deployed in the complete localize-and-classify pipeline. These crops are expected to be either inaccurate or incorrect altogether.

Towards this end, we added an extra non-character class corresponding to bounding boxes containing no character or damaged characters. The non-character images  simulates false positives detected by the previous localization stage. 

For all characters, we provide their ground truth bounding box coordinates, except for the non-character class, which has a bounding box size equal to the mean character size of the seal image to which it belongs. In summary, 150 non-character images are randomly generated at each iteration of the K fold cross-validation processes and added to the character image set. 

Table \ref{tab:number_characters} shows the number of seals used for the localization task. Good quality obverse images are added for augmenting the data.
For the classification task,  we provide the number of characters used for this task. It can be noted that these characters are cropped from the seal images used for localization. Folds at character level (classification task) follow the folds at seal level (localization task). This ensures that training and test characters always belong to  distinct seals (see Section \ref{sec:protocol}).
%
\begin{table*}[h!]
     \caption{Number of seal images for the character localization task (top) and number of characters for the character classification task (bottom). Samples will be divided into training/test data according to a cross-validation framework. }
    \centering
    \begin{tabular}{||c|c c c c  ||}
    \hline
        &  reverse & obverse   & non chars & total  \\
      \hline
      Character Localization (YOLOs v5) & 66 & 36 & - & 102 \\
      \hline
      Character Classification (ResNet18) & 1857 & 456 & 150 & 2463 \\
 \hline
    \end{tabular}
    \label{tab:number_characters}
\end{table*}

\subsection{Text transcription}
\label{sec:text_transcription}

Text lines are extracted from bounding boxes and label predictions using a Hough-based approach described in \cite{likforman95}. This approach consists in detecting the most likely lines of text (the ones that intersect the most bounding boxes) following a hypothesis-validation strategy which is iteratively activated until the end of the segmentation is reached. At each stage of the process, the best text-line hypothesis is generated in the Hough domain. Afterwards, the line's validity is checked in the image domain using proximity criteria.
Finally, a diplomatic transcription is obtained by combining the previously detected lines with character labels (lines of the transcription correspond to those of the input image).

\section{Experiments}\label{sec:experiments}
We describe in the following, the evaluation protocols for each task. We then evaluate each task in isolation, and in combination for the entire pipeline.

\subsection{Evaluation protocol}
\label{sec:protocol}
%
%
We adopt a cross validation framework with K folds,
i.e. we randomly draw K folds out of the annotated data and in turn we leave out one fold for testing while we train on the remaining K-1 folds.
All the results presented below are relative to the reverse side of seals and they all represent the mean of the K folds performance.
For the localisation task, the folds include seal images and  for the classification task, characters. To avoid including characters from the same seal  in training and testing folds, the same folds used for localization, are used for classification. The training character images are cropped from the training seal image folds, and the same for the testing fold.
Since the characters on the reverse images include rare letters, and our dataset is scarce, we have augmented our data by adding to the training sets of both tasks, a few observe seal images and characters. 
%
%

In Section \ref{sec:eval_recognition}, 
the  classification task is evaluated from ground truth bounding boxes.
However when we will evaluate  the complete pipeline (see Section \ref{sec:eval_transcription}), we will consider the  bounding boxes provided by the first CNN (YOLO, character localizer).
\subsection{Character localization}\label{sec:eval_localization}
\begin{figure*}[h!]
    \centering
    \includegraphics[width=0.8\linewidth]{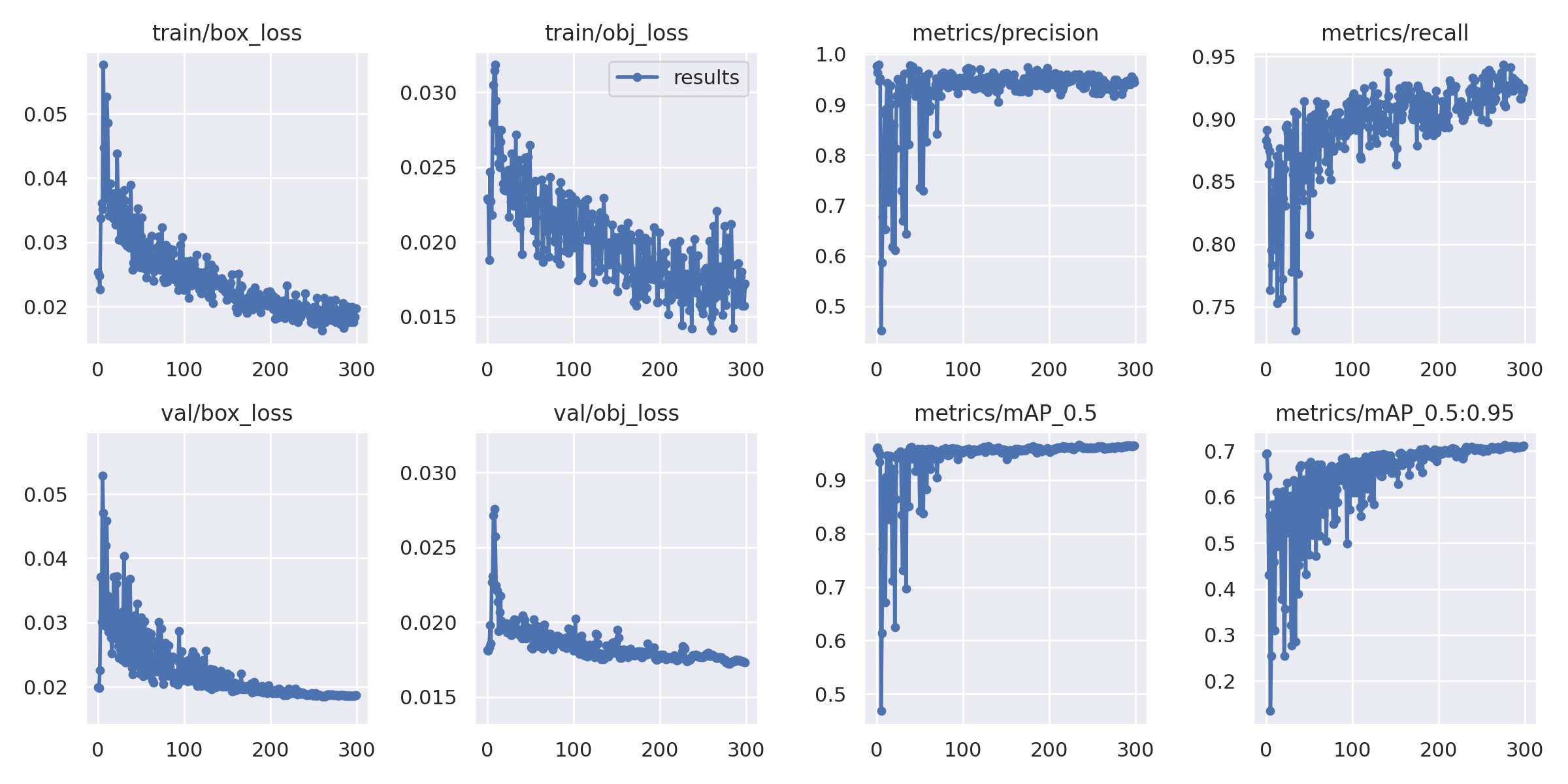}
    \caption{Training (top) and validation (bottom) loss curves (two columns on the left), and evolution of recall, precision and mAP metrics according to the number of epochs (two columns on the right).
    }
    \label{fig:results_yolo_onecls}
\end{figure*}

%
To validate the YOLO system used as a character localizer, we use cross-validation with $K=10$ folds. 
Figure \ref{fig:results_yolo_onecls}  shows the evolution of the training and validation losses (two columns on the left) and the metrics on a validation set (two columns on the right) over epochs. The validation set includes 15\% of the training data.
%
%
%
%
Figure \ref{fig:results_yolo_onecls} shows that the mAP@[0.5:0.95] reaches 0.7 after 300 epochs 
which shows that our model is confident when predicting bounding boxes. In fact, a high value of this metric indicates that even for high values of the IoU threshold (for instance $IoU_{t} = 0.9$ or $0.95$), the model achieves decent results.
For obtaining the cross-validation performance, we cycle over the folds.  
For each iteration of the cross validation, the best model on the validation set is recorded and tested on the test fold. Table \ref{table:cv_detection} provides Cross validation performance on reverse images for three popular metrics (recall, precision, mean average precision mAP).

%
\begin{table}[h!]
\caption{ Character Localization evaluation on reverse seals.
Recall, precision and mAP metrics (in \%) obtained by cross validating over K = 10 folds. The value of the IOU threshold equals 0.5.} 
 
\centering
\begin{tabular}{||c | c | c c ||} 
 \hline
 Precision & Recall & \multicolumn{2}{c ||}{mAP} \\
 \hline
  &  & @0.5 & @[0.5:0.95] \\
 \hline
 \hline
 93.74 & 89.80 & 94.40 & 65.78 \\
 \hline
 \end{tabular}
\label{table:cv_detection}
\end{table}

%

\subsection{Character classification  }\label{sec:eval_recognition}

Next, we evaluate the performance of the ResNet18 network in charge of classifying the characters localized by the YOLO character detector.
We decided to train our model on ground truth crops rather than on the deep network detector predictions because they might be inaccurate compared to the ground truths and might introduce a bias in our dataset. We also add padding around each crop in the training set to make the network more robust for predicting on new data.
The results in Table \ref{table:cv_classification} are  relative to the ResNet18 network evaluated on ground truth crops for the  20 most represented classes (which are composed at least of 50 samples) plus the non-character class defined in Section \ref{sec:Resnet}.

Table \ref{table:cv_classification} provides cross-validation accuracies and F1-scores for the classification task computed on the reverse side of test images.
We trained the classification model from pretrain weights learned on the ImageNet dataset (\cite{ImageNet}). We used some common data augmentation such as affine transformation and change of colour. For each iteration of the cross-validation procedure, we train a model for 100 epochs with early stopping. For each training, the best model in terms of Top-1 accuracy and F1-score on the validation set (15\% of the training set)  is recorded and tested on the test fold.

\begin{table}[h!]
\caption{Character classification accuracies and F1-Score obtained by cross-validation for $K = 10$ folds. All metrics are provided in percentages (\%) }.
\centering
\begin{tabular}{||c c c | c||} 
 \hline
 Top-1 acc.  & Top-2 acc. & Top-3  acc. & F1 Score \\ [0.5ex] 
 \hline
 92.12 & 96.10 & 97.37 & 91.65 \\
 \hline
 \end{tabular}
\label{table:cv_classification}
\end{table}

\begin{figure}[h!]
    \centering
    \includegraphics[width=0.8\linewidth]{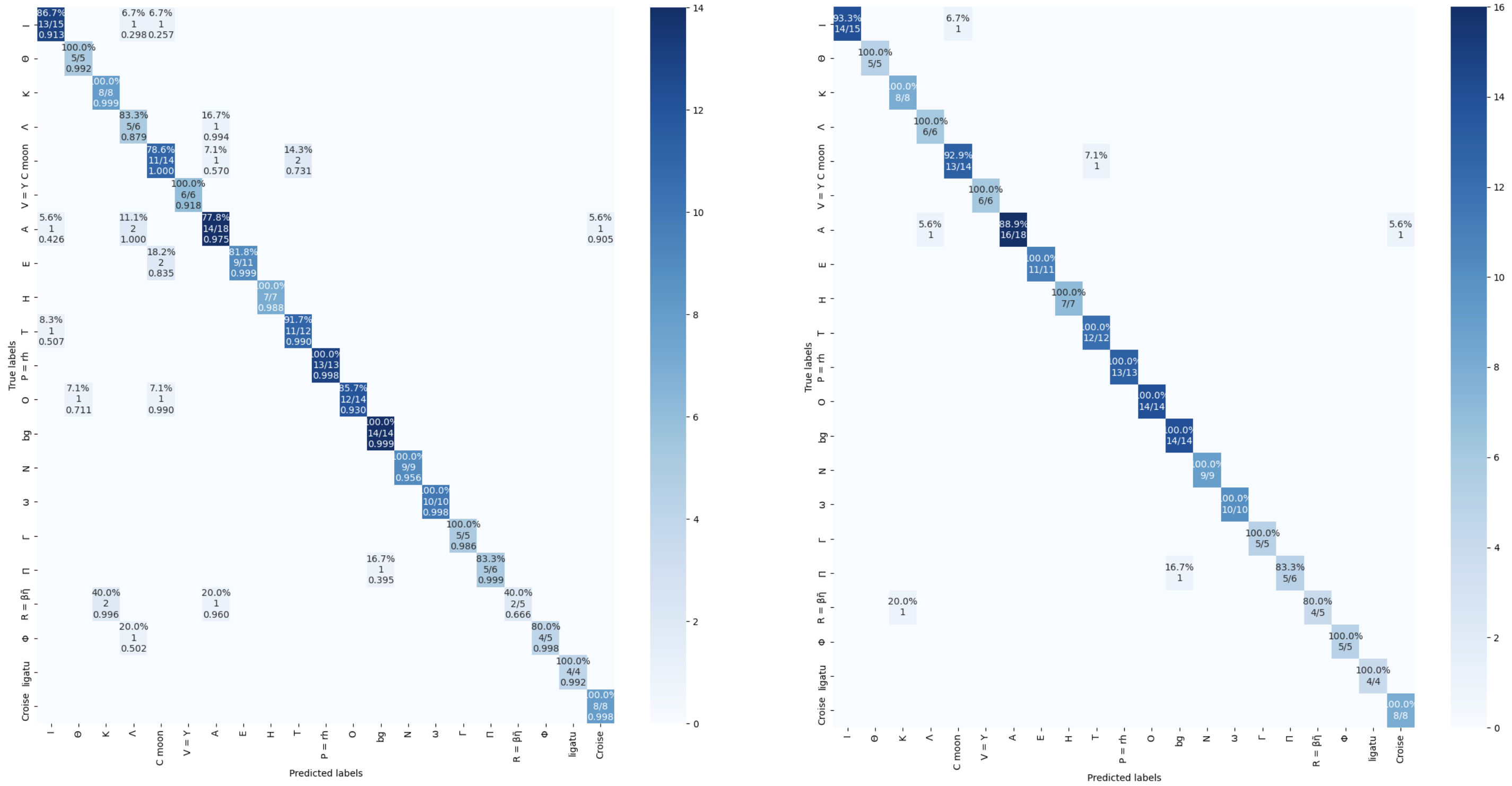}
    \caption{Top-1 (left) and Top-3 (right) confusion matrices for the character classification task, on one test fold. We consider the 20 most represented (at least 50 samples) character classes, plus one pseudo-class for false positives. 
    }
    \label{fig:conf_mat_resnet}
\end{figure}

Figure \ref{fig:conf_mat_resnet} shows the confusion matrices (Top-1 on the left and Top-3 on the right) obtained after one single training of the cross-validation procedure.
The Top-1 confusion matrix (left) shows that the model achieves decent results but still makes some mistakes. The Top-3 confusion matrix is almost diagonal, it indicates that the model has understood well how to dissociate the classes and makes only some mistakes on the hard examples.

%
\subsection{Transcription Evaluation }\label{sec:eval_transcription}
In the previous section, the pipeline classification performance has been evaluated from ground truth bounding boxes. However, bounding boxes predicted by the YOLO character detector may be inaccurate and thus differ from the ground truth. Several cases explain such difference:
\begin{itemize}
    \item the bounding boxes (ground truth/predicted) overlap but are not identical (their IOU is greater than a threshold (f.i. 0.5) )
    \item a bounding box is predicted at a location where there is no character (False positive)
    \item a character has been missed (no bounding box predicted, or IOU less than 0.5)
\end{itemize}
In the first case, even if bounding boxes are not identical, the ResNet18 character classifier could recognize the character cropped with the predicted box. 
Thus we evaluate the pipeline through the  predicted transcription, the sequence of characters
output, whatever their precise location.
However, the output of the application of ResNet18 to all the bounding boxes detected by YOLO is a set of class score distributions. To obtain a diplomatic transcription, these characters are first grouped into text-lines (see Section \ref{sec:text_transcription}). Then, they are ordered within each line. 
%

\begin{figure}[h!]
    \centering
    \includegraphics[width=0.8\linewidth]{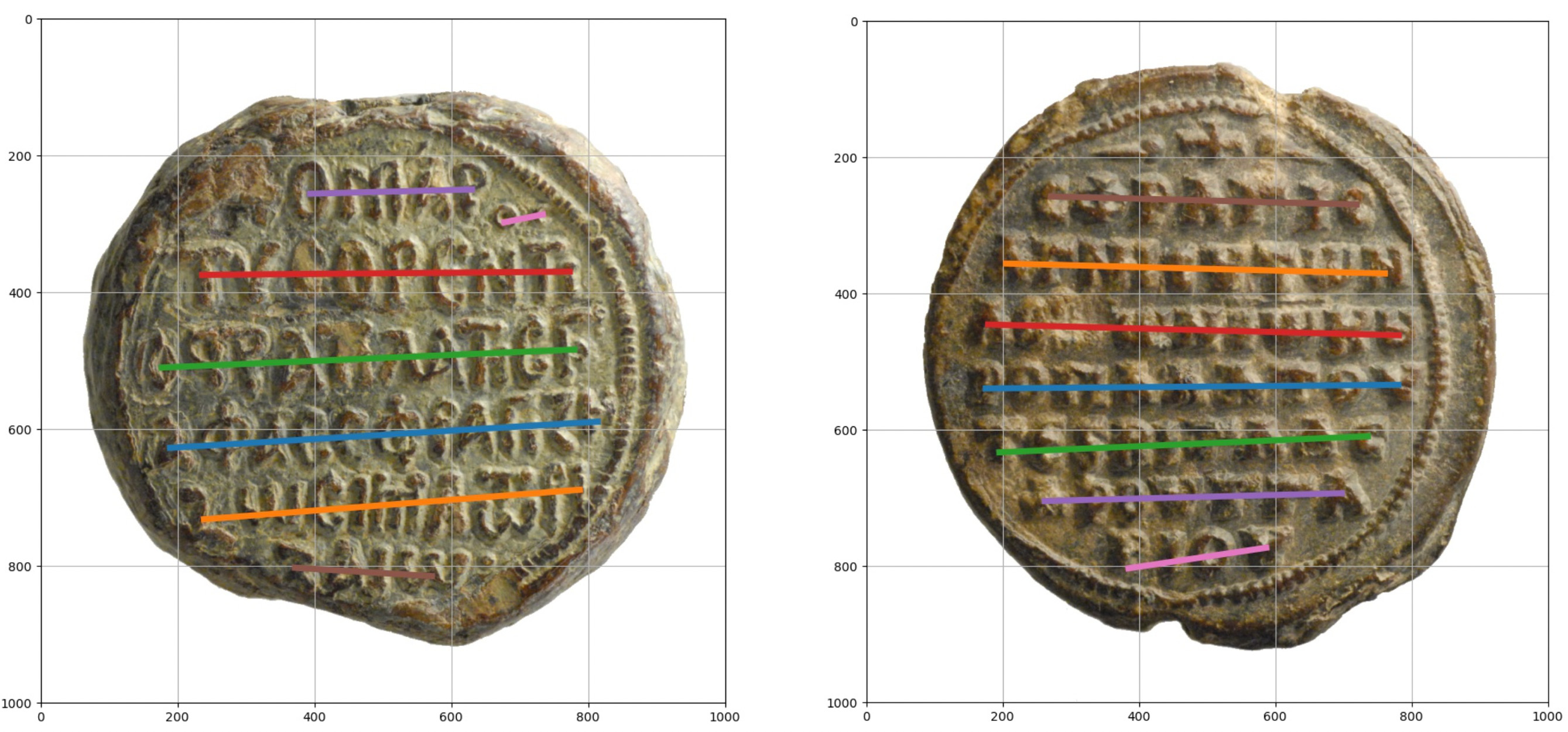}
    \caption{Text lines extracted from sample seal images.}
    \label{fig:hough}
    \end{figure}

\begin{figure}[h!]
    \centering
    \includegraphics[width=0.7\linewidth]{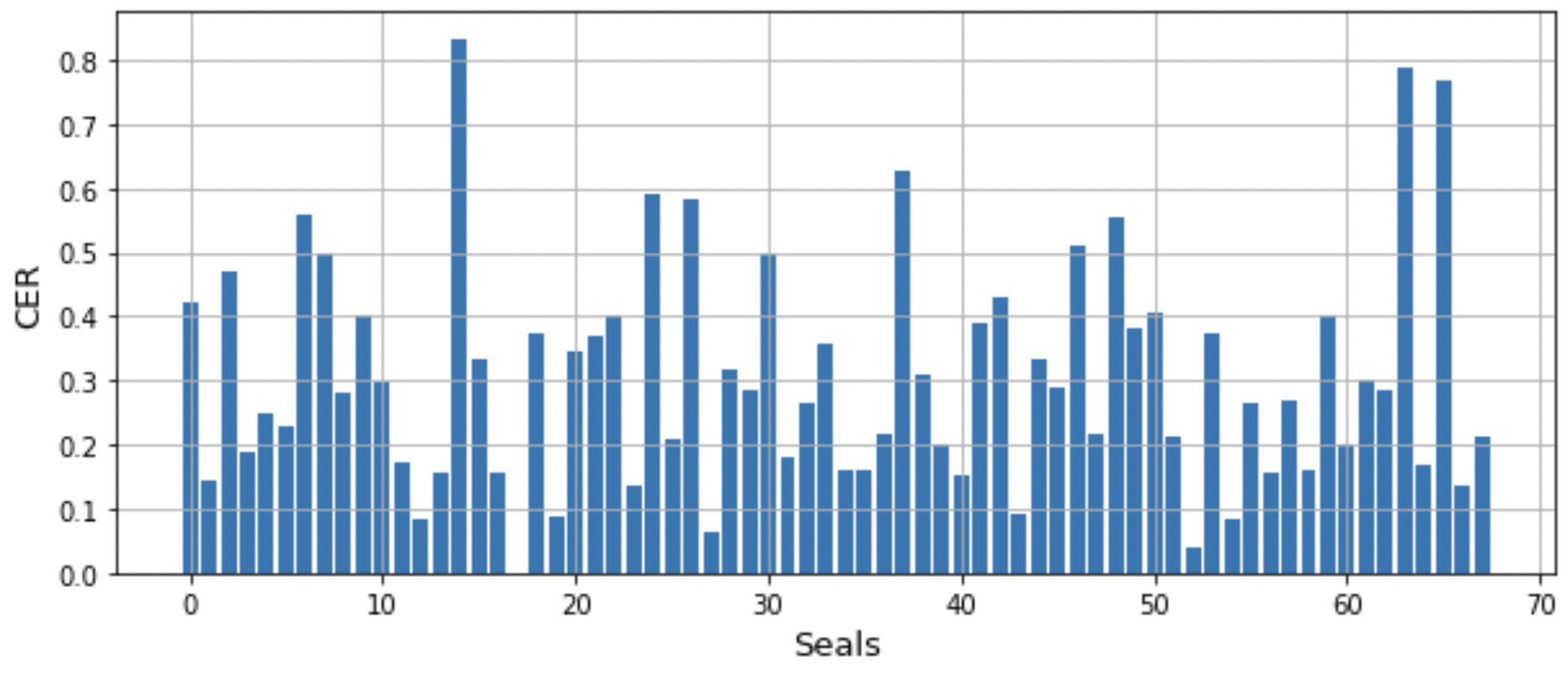}
    \caption{Character Error Rates obtained for each seal reverse image for the complete pipeline.}
    \label{fig:CER}
\end{figure}

%
Figure \ref{fig:hough} shows sample seals and their extracted text lines.
To compare the predicted character sequence of one seal with its ground truth sequence, we adopt the CER metric (Character Error Rate). This metric relies on the Levenshtein distance. It is equal to the minimum number of characters that must be deleted, inserted or substituted to move from the predicted to the ground truth sequence.
To obtain the CER, we divide this distance by the number of characters in the ground truth sequence.
%
$$ CER=\frac{S+D+I}{N} $$

where S is the number of character substitutions, I the number of insertions, and D the number of deletions. N is the  number of characters in the seal ground truth sequence.
Following the K-fold cross-validation framework, we compute for each seal of the testing fold, its CER, and average over the seals to obtain the CER associated to this fold.  The cross-validation CER is obtained by averaging over the folds (see Table \ref{table:cross_val_edition}). 
The Cross-validation CER equals 0.31, 
which proves the efficiency of the proposed approach for seals' written artifacts. 
In Figure \ref{fig:CER}, we show all seal image CER values. High values of CER correspond to hard seal examples, either damaged or containing illegible characters. In this case, the detector results are less accurate (bounding boxes that do not cover all characters or the presence of false positives). The smallest CER values correspond mostly to classification errors (substitution between two closely shaped characters).
\begin{table*}
\centering
\caption{Cross-validation results (K=10) for the complete pipeline (transcription task).  We provide the CER (Character Error Rate) for each fold, and the overall performance.  }\label{table:cross_val_edition}
\begin{tabular}{|l|l|l|l|l |l | l |l |l  | l | l  | l |} 
\hline
 Fold & 1 & 2 & 3 & 4 & 5 & 6  & 7 & 8 & 9 & 10 & \textbf{overall} \\
 \hline
 CER  & 0.32 & 0.27 & 0.30 & 0.33 & 0.29 & 0.29 & 0.34 & 0.25 &  0.24 & 0.39 & \textbf{0.31}\\

\hline 
\end{tabular}
\end{table*} 

\begin{figure*}[h!]
    \centering
    \includegraphics[width=1\linewidth]
{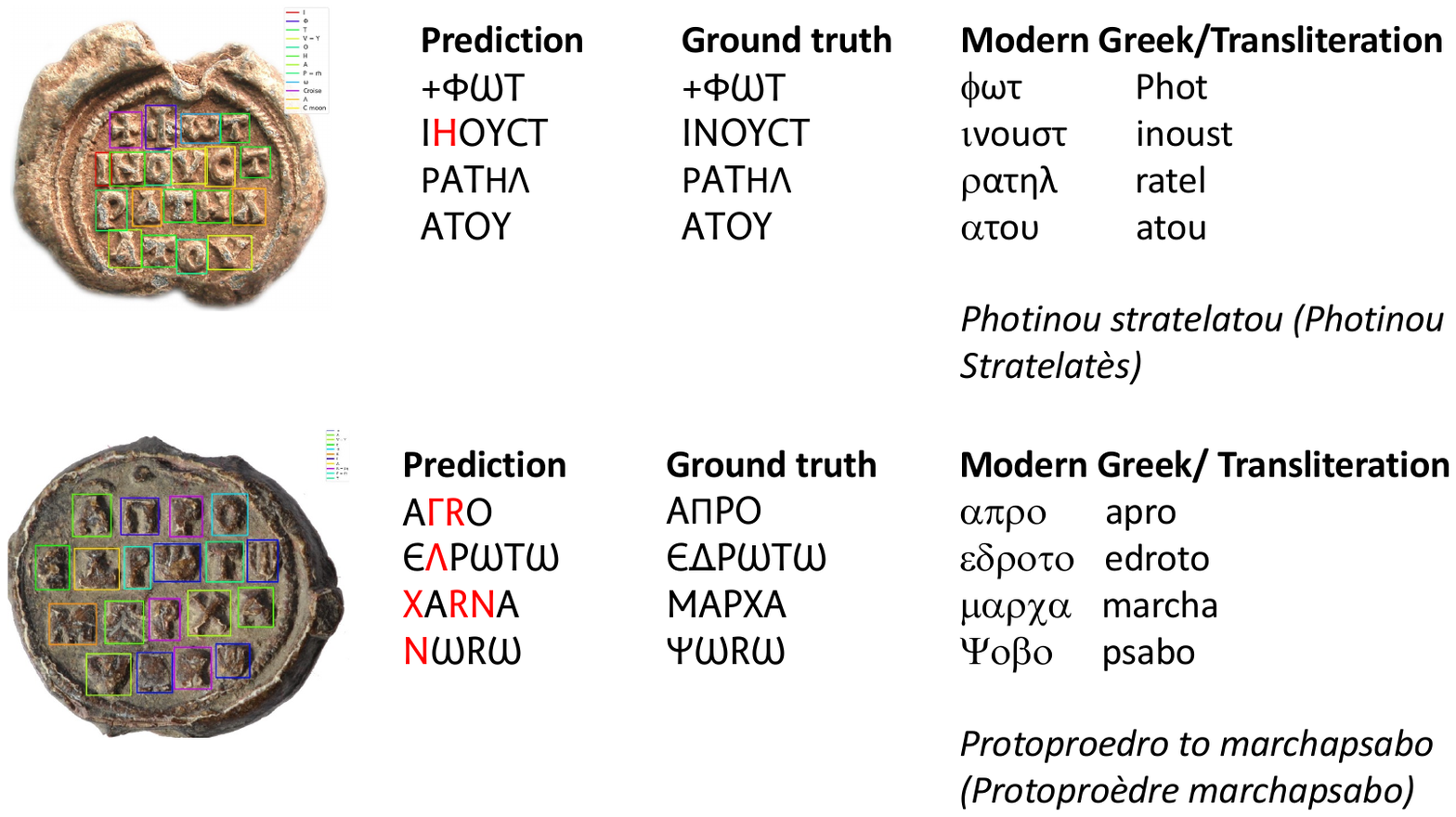}
    \caption{Transcriptions obtained for two samples seals. Characters in red denote an error. 
    }
    \label{fig:res_transcription}
\end{figure*}
Figure \ref{fig:res_transcription} shows the transcriptions obtained for two sample seals. In both seals, characters have been correctly detected by the deep localizer network. Characters have been almost perfectly recognized for the top seal. There are a few errors in  the second one. Errors are due to confusions between characters. It can be noticed that characters of the second seal are much less legible, compared  to the  first seal. 
In this figure, we have added the ground truth in modern Greek,  its  transliteration, and the underlying text provided by an expert. The first $\alpha $ in the second seal is the abbreviation of word 'proto'.
A \textit{Stratelatès} is an army general. A \textit{Protoproèdre} is a byzantine dignitary (X-XI\textsuperscript{th} centuries), and Marchapsabo, the name of a well-known family. 

\begin{figure}[h!]
    \centering
    \includegraphics[width=0.6\linewidth]
{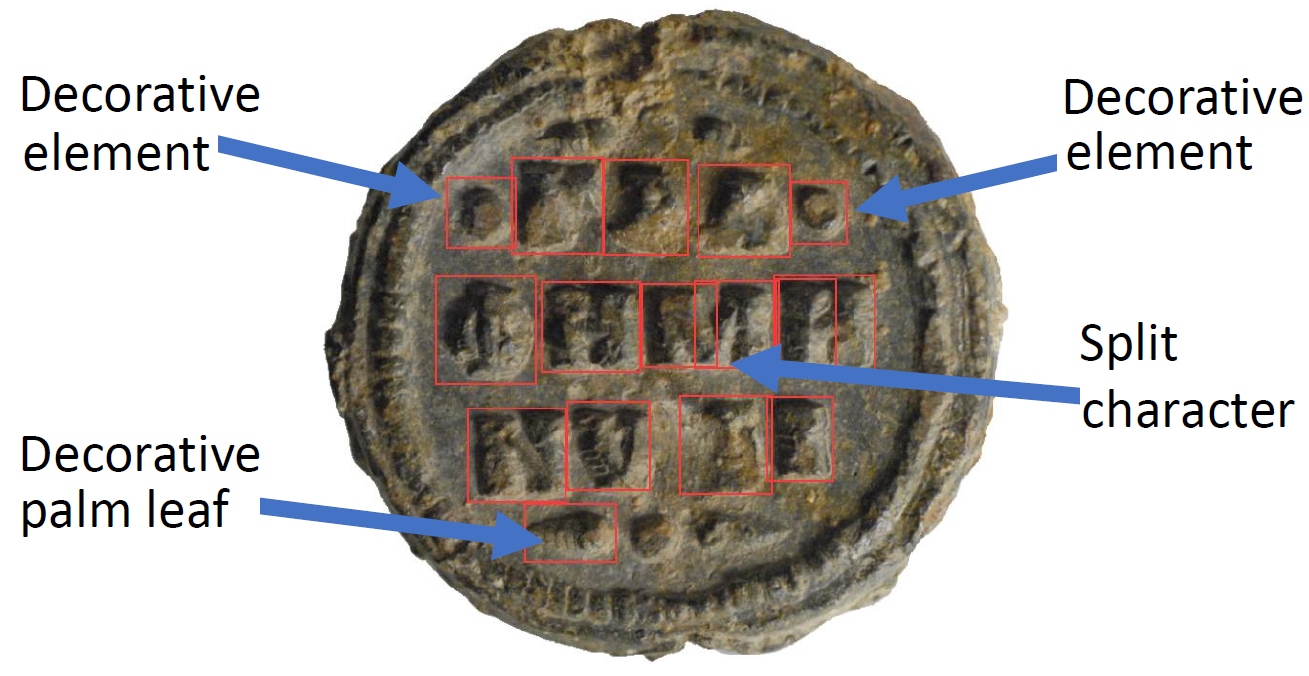}
    \caption{Hard case samples. Localization errors due to decorative elements. Large characters such as 'M' may be split. 
    }
    \label{fig:res_hard_case}
\end{figure}

    %
    Figure \ref{fig:res_hard_case} shows typical difficult  cases.  Decorative elements (o-shaped, leaves) can be  seen as characters. The large letter 'M'  is split into its compound elements.

\section{Discussion}
An overall character error rate of about 30\% cannot be directly compared with high accuracies reported on standard datasets (such as the digit MNIST dataset (\cite{mnist})) with deep networks. Reported errors consist of confusion between classes only. Moreover, such sets include thousands of samples, so there is no scarce data problem. Characters are also well-contrasted and correctly cropped; thus, there are no occluded or missing strokes.  
In contrast, the character error rate in our overall pipeline reports several errors from the predicted character sequence. Errors include confusions but also false positive and false negative crops. Confusion errors may be due to non-perfect crops with missing strokes, which is not the case with standard character recognition datasets.  

A fairer comparison may be made in the numismatic domain. The CoinNUMs\_pcgs\_m and CoinNUMs\_pcgs\_a datasets collected by \cite{pan17} include digit images manually or automatically cropped from coin images, respectively. The foreground and background share the same hue (as in our case), and there may be crop errors in the automatic setting. Performance for manually cropped characters is greater than for automatically cropped ones (43.2 \%  accuracy versus 29.4\% using the mean over classes of the per-class accuracy ).
%
This can be explained by the fact that automatic cropped characters include occluded strokes, or are not centered. For manually cropped characters, the relatively low performance can be explained by the small size of the digit dataset, and the use of a too basic neural network-based classifier. Using the same metric, our average per-class accuracy is equal to 89\% considering the most frequent classes (20 classes)  and the non-character class. Our performance thus compares favorably, and we also take into account errors due to false positives.
\footnote{Code and  trained neural networks weights will be released upon article publication.}
\section{Conclusion and perspectives}
\label{sec:conclusion}

In this work, we have proposed a two-stages approach for reading Byzantine seals. A first deep neural network localizes the characters within the seal, whereas a second network classifies the previously localized characters. A character-level diplomatic transcription is then obtained by reordering characters into text lines. This transcription is the first step towards language processing to extract words, named entities, and recover the underlying text.

However, our architecture presents several constraints that we are planning to overcome.
\\
First, our system relies on a two-stages (localize and classify) approach since we do not have enough annotated images to train a single architecture performing joint character localization and classification. Towards this end, we have achieved encouraging preliminary results by boosting our training set with computer-generated seal images. Whether synthetic images can replace real images in training is the object of our current endeavors.
\\
Second, in this work, we have restrained ourselves from transcribing the reverse side of the seals where the text is oriented on horizontal lines. The obverse side may include text oriented on a curve and is often cluttered with images, hindering the transcriptions. Our future research also includes transcribing the obverse side by properly detecting the character's orientation and 
location within the seal.
\\
Third and last, inscriptions on seals often correspond to abbreviated texts that the expert readers of that time were supposed to be able to decipher. Furthermore, seals often present damages that result in the loss of one or more consecutive letters. Natural language processing approaches may help reveal the original underlying text: for instance, in \cite{ithaca22}, the damaged text is completed by a deep neural approach. 

\subparagraph{Funding: This research was funded by the ANR French National Research Agency, grant number ANR-21-CE38-0001}.

\bibliographystyle{cas-model2-names}

\bibliography{cas-refs}















\end{document}